\documentclass[final,journal]{IEEEtran}

\usepackage{cite}

\ifCLASSINFOpdf
   \usepackage[pdftex]{graphicx}
\else
\fi
\usepackage{array}

\ifCLASSOPTIONcaptionsoff
  \usepackage[nomarkers]{endfloat}
 \let\MYoriglatexcaption\caption
 \renewcommand{\caption}[2][\relax]{\MYoriglatexcaption[#2]{#2}}
\fi
\begin{document}
%
\title{It's the Journey Not the Destination: Building Genetic Algorithms Practitioners Can Trust}


\author{\IEEEauthorblockN{Jakub Vincalek\IEEEauthorrefmark{1},
Sean Walton\IEEEauthorrefmark{1},
Ben Evans\IEEEauthorrefmark{2}
}
\IEEEauthorblockA{\IEEEauthorrefmark{1}College of Science, Swansea University, Swansea, UK}
\IEEEauthorblockA{\IEEEauthorrefmark{2}College of Engineering, Swansea University, Swansea, UK}}

%



\IEEEtitleabstractindextext{%
\begin{abstract}
Genetic algorithms have been developed for decades by researchers in academia and perform well in engineering applications, yet their uptake in industry remains limited. In order to understand why this is the case, the opinions of users of engineering design tools were gathered. The results from a survey showing the attitudes of engineers and students with design experience with respect to optimisation algorithms are presented. A survey was designed to answer two research questions: \emph{To what extent is there a pre-existing sentiment (negative or positive) among students, engineers, and managers towards genetic algorithm-based design?} and \emph{What are the requirements of practitioners with regards to design optimisation and the design optimisation process?} A total of 23 participants (N = 23) took part in the 3-part mixed methods survey. Thematic analysis was conducted on the open-ended questions. A common thread throughout participants responses is that there is a question of trust towards genetic algorithms within industry. Perhaps surprising is that the key to gaining this trust is not producing good results, but creating algorithms which explain the process they take in reaching a result. Participants have expressed a desire to continue to remain in the design loop. This is at odds with the motivation of a portion of the genetic algorithms community of removing humans from the loop. It is clear we need to take a different approach to increase industrial uptake. Based on this, the following recommendations have been made to increase their use in industry: an increase of transparency and explainability of genetic algorithms, an increased focus on user experience, better communication between developers and engineers, and visualising algorithm behaviour.

\end{abstract}}


\maketitle

\IEEEdisplaynontitleabstractindextext


\IEEEpeerreviewmaketitle

\section{Introduction}
%
%
%
%

 


\IEEEPARstart{T}{his} paper aims to provide insights for developers and researchers of genetic algorithms and engineering design tools. These insights include the attitudes that engineers and students have towards genetic algorithms. The current barriers they face in the design process were explored. A total of 23 participants were recruited for a mixed methods survey. Their opinions were analysed thematically as a means of establishing their preferences with respect to engineering design tools and genetic algorithms. This paper provides a number of recommendations for developers and researchers. The contribution of this paper to the domain of evolutionary computing is establishing best practices with respect to the development of genetic algorithms in the context of design engineering.

\subsection{Genetic Algorithms}

Genetic algorithms (GAs) are nature-inspired algorithms. They use the process of natural selection through selection, crossover and mutation to search for an optimal solution to a problem. The solutions are driven to optima by a fitness function which dictates which traits are selected for the next generation. Some common examples of GAs are Particle Swarm Optimisation \cite{trelea2003particle}, Artificial Bee Colony \cite{karaboga2007powerful}, Ant Colony Optimisation \cite{dorigo2006ant}, Firefly Algorithm \cite{yang2013firefly} and Cuckoo Search Algorithm \cite{gandomi2013cuckoo}. Examples of natural phenomenon GAs are Colliding Bodies Optimisation \cite{kaveh2014colliding} and Gravitational Search Algorithm \cite{rashedi2009gsa}.

\subsubsection{Applications}
GAs have proved to be a powerful tool for optimisation \cite{wang1997using, oduguwa2002bi, mcgookin2000ship}. Their application ranges from very simple mathematical formulae to complex systems with many interdependencies. The design of a simple cantilever beam \cite{mirafzal2016optimizing} and elaborate aerodynamic design \cite{evansaerodynamic} showcase the range of applications. To measure the effectiveness of an algorithm, they are often tested against standard benchmark functions such as the Ackley Function \cite{ballester2004effective} and constrained engineering problems like a welded beam or a pressurised vessel \cite{zhang2018queuing}. 

Multi-objective optimisation (MO) problems are particularly well suited to GAs. Many engineering problems can be classified as MO; these designs involve many different trade-offs. It is often challenging for human designers to determine an optimal solution without the aid of a computer-based system. One example of an MO problem where a GA was used successfully is presented by Tawhid \& Savsani \cite{tawhid2018novel} who evaluated their Artificial Algae Algorithm on 20 different benchmark problems to show its suitability for engineering design optimisation problems. An industrial application of MO engineering can be found in aerospace, specifically the design of geostationary satellites. Customer demands for higher bandwidth capabilities and longer lifespans \cite{airbus2020tele} are at odds with the many constraints in place. These constraints include
\begin{itemize}
\item Higher power requirements;
\item Thermal considerations, including hot and cold zones;
\item The ability to fit in a rocket fairing; and
\item The ability to withstand vibrations during launch.
\end{itemize}
Each of these variables need to be considered throughout the design lifecycle of the satellite. To illustrate the usefulness of GAs in this context, Berrezzoug \emph{et al.} \cite{berrezzoug2019interactive} proposed a method which applies a Gravitational Search Algorithm to the design of a satellite by considering the many design variables. The approach proved useful for determining optimal trade-offs between the variables.

Optimisation problems do not need to be strictly product design based either. The layout of a construction site is another kind of MO problem. Offices, equipment and warehouses are among the factors which influence the layout. The objective for any layout is to maximise the time spent on value adding activities while decreasing wasteful activities like transportation or waiting. In addition to this, the safety of the construction workers and engineers cannot be compromised. One attempt at this was done by Kaveh \emph{et al.} \cite{kaveh2016construction} who demonstrate the application of a Colliding Bodies Optimisation algorithm to a construction site layout. Their solution demonstrates that a problem with \emph{n!} combinations can be effectively solved with a GA.

The optimisation problems presented in this section are bounded by the constraints put on them by external factors, such as launch conditions for a satellite. In the following section, the constraints that designers face, such as resources and time, will be explored. These factors present barriers for the designers when attempting to optimise their designs. 

\subsubsection{The Relationship Between Humans and Algorithms} 

Rather than acting as autonomous agents, GAs can be integrated in the design process to enhance the capabilities of human designers. In recent years, a growing trend of combining the expertise of designers with algorithms has been taking place. For example, Ant Colony Optimisation was applied to the design process of the layout of train tracks \cite{flurl2016energy}. The layout can be optimised to reduce the energy consumption of the trains as they travel the tracks. Furthermore, Flurl \emph{et al.}  \cite{flurl2016energy} propose a method to allow design engineers to see the effects of their changes on energy consumption in real-time. This method takes advantage of the algorithm's ability to find an optimal solution while allowing the designer to stay in control of the track layout.

The push for user-centric design is evident in engineering as well as other disciplines such as architecture. In a 2017 survey of 165 architects, 91\% of respondents would like to see an inclusion of a ``human-in-the-loop" approach with regard to the software and technologies they use \cite{cichocka2017optimization}. Furthermore, 93\% of respondents wanted to be able to understand the underlying principles of the algorithm while just over half (54\%) would like full control over the process. In an attempt to bring this to fruition, Berseth \emph{et al.} \cite{berseth2019interactive} propose an interactive Computer-Aided Design (CAD) optimisation program. It engages with architects by analysing their designs and optimising for a number of variables such as ``open space in passages, aesthetic relationships, or building codes." This is akin to the MO problems encountered in aerospace engineering applications such as satellites, but also in the development of rockets. In fact, a user-centric approach was adopted by engineers at NASA \cite{burnell2020gpkit} and allowed engineers to evaluate the interdependencies that different systems had to each other with respect to the mission objective. 

Other approaches attempt to rely too heavily on algorithms alone and exclude human intervention. This approach leads to systems either whose complexity is increased or explainability is decreased. Explainability is the ability for a developer or user to understand a model's behaviour \cite{pasquale2017toward}. Two examples of approaches which rely too heavily on algorithms have been shown when trying to solve MO problems. The first is a Symbiotic Organisms Search algorithm which was developed by Ustun \emph{et al.} \cite{ustun2020symbiotic}. It is a very recent algorithm that may perform better than other standard GA for certain MO problems, but falls victim to the No Free Lunch Theory \cite{wolpert1995no}. The No Free Lunch Theory states that for all problems in which a supposed \emph{Algorithm X} outperforms \emph{Algorithm Y}, there are an equal amount of problems in which \emph{Algorithm Y} outperforms \emph{Algorithm X}. In short, there does not exist an algorithm that will be the best choice in all scenarios. The development and tuning of GAs is itself an optimisation problem and GAs are developed on a case-by-case basis with their applicability to different situations being very limited. The second example of an over-reliance on algorithms has to do with eliminating the consideration for external factors on the optimisation process. Fleck \emph{et al.} \cite{fleck2018novel} discuss a method to reduce an MO problem down to a single value. This is a problem for any real-world engineering application. Take for example the previously mentioned satellite problem. There are many design variables to consider and their dependencies must be investigated in depth. The lack of explainability would be difficult to justify in satellite development. A need for explainability was also noted by Burnell \emph{et al.} \cite{burnell2020gpkit} when incorporating their approach at NASA.

By combining human expertise with algorithms, designs that are both optimal and adhere to all design constraints can be produced. This is demonstrated by Guo \emph{et al.} \cite{guo2019novel} for bolt supporting networks. Users could select from a range of solutions which were fed back to the algorithm to develop a solution that was strong, cost-effective and manufacturable. Oulasvirta \cite{oulasvirta2017user} considered user-centric design when applied to User Interface (UI). One of the key points raised with the approach was that the ``designer can steer and redefine the tasks intermittently as results stream in."  Thus, an optimum between designer and computer also exists; a region in which the algorithm is able to perform the necessary calculations while the human designer can focus on producing a result in line with their vision. The real-time approach is adopted by Umetani \& Bickel \cite{umetani2018learning} and is applied to automobile aerodynamics. The changes made by users and the effects those have on the drag characteristics of the vehicle can immediately be fed back to the designer. Incorporating humans in the optimisation process indirectly involves designers from outside of the design process. This is because engineers do not work in isolation, but rather as a team. Adjoul \emph{et al.} \cite{adjoul2019algorithmic} observed that optimisations in the production phase required experts from both production and design to work together to create an optimal product. The entire life cycle of a product must be considered when dealing with optimisations. An instance of this is probed by Kang \emph{et al.} \cite{kang2019form+} while considering the trade-offs faced by customers when purchasing a new car. Their techniques ``could readily lead to crowdsourced, real-time, manufacturer-feasible design optimization [sic]," which includes customers, designers and engineers.

The integration of designers in the design loop has been shown to be possible. The results, however, have been focused on the quantitative aspect of the application. An in-depth thematic analysis of the opinions of practitioners has not been evaluated in all the previously mentioned papers, which this paper contributes to the field.

\subsection{Research Questions}

Academic literature has shown that the use of GAs can add value in industry. However, the best way to do so remains unclear. Gaining a familiarity with the way designers work can be used to develop a better understanding of why the uptake remains low. The following research questions were posed:
\begin{enumerate}
\item To what extent is there a pre-existing sentiment (negative or positive) among practitioners towards genetic algorithm-based design?
\item What are the requirements of students, engineers and managers with regards to design optimisation and the design optimisation process?
\end{enumerate}

\section{Method}

\subsection{Survey}

The survey consisted of 3 parts. Part 1 was used to gather non-identifying traits about the respondent such as their experience in the field and their roles. Part 2 gauged the respondents prior knowledge of the domain as well as their preferred characteristics for design tools. Part 3 of the survey had 6 optional, open-ended questions aimed at determining the current state of a design process and the respondents attitudes towards algorithmic design aids. There was no time limit for responses, however, 20 minutes was suggested as an estimate of time for completion.

The analysis of the results have been broken down into three parts to reflect the structure of the survey. Thematic analysis has been used for Part 3, while Parts 1 and 2 can be analysed quantitatively. A total of 23 participants (N = 23) answered the survey over the course of 9 days. All participants answered all questions in Parts 1 and 2. Some respondents chose not to answer some or all questions in Part 3. The total number of responses per question in that Part are given along with the result.

The survey was created electronically using Google Forms. A link to a survey was posted on various social media sites (Twitter, LinkedIn, Facebook, Reddit) to gather responses; emails were also sent directly to those with known prior domain experience. Ethics approval for this study was granted by the Swansea University College of Science Ethics Committee (SU-Ethics-Student-110620/2921). 

\subsection{Materials}

Participants were asked to categorise themselves based on their current roles. The sum of the responses exceeds 23 as some participants may feel that more than one category applies to them. More than half (N = 13; 57\%) of the respondents considered themselves as engineers, with roughly a third (N = 8; 34\%) specifically selecting ``Design Engineer." Some participants (N = 5; 22\%) selected more than one descriptor.

\begin{table}[htp]
\caption{A breakdown of roles of participants.}
\begin{center}
\begin{tabular}{| l | r |}
\hline
Role & Count \\ \hline
Student & 9 \\
Researcher & 1 \\
Design Engineer & 8 \\
Other Engineer & 7 \\
Manager & 3 \\
Retired & 0 \\ \hline
\end{tabular}
\end{center}
\label{roles}
\end{table}%

Participants were asked to state their highest level of education. All participants but one (N = 22; 96\%) hold either an undergraduate or postgraduate degree. Slightly under half (N = 10; 43\%) of participants hold a postgraduate degree. The specific degree was not probed.

\begin{table}[htp]
\caption{A breakdown of education levels of participants.}
\begin{center}
\begin{tabular}{| l | r |}
\hline
Education & Count \\ \hline
Secondary School & 1 \\
Undergraduate & 12 \\
Postgraduate & 10 \\ \hline
\end{tabular}
\end{center}
\label{education}
\end{table}%

In addition to their education, respondents were asked to state their length of experience in the field of engineering design as well as their self-declared proficiency. The proficiency was declared on a 6-point Likert scale (from ``Not Proficient Whatsoever" to ``Extremely Proficient"). Of all participants, just under a third (N = 7; 30\%) indicated they had more than 10 years of experience in this field. Moreover, most (N = 19; 83\%) participants indicated they are at least ``proficient" in engineering design. Over half (N = 10; 43\%) of those 19 would say that they are very proficient while one (N = 1;  4\%) participant indicated extreme proficiency in engineering design.

\begin{table}[htp]
\caption{A breakdown of experience levels of participants.}
\begin{center}
\begin{tabular}{| l | r |}
\hline
Experience & Count \\ \hline
Less than 1 year & 2 \\
1 to 2 years & 7 \\
3 to 5 years & 6 \\
6 to 10 years & 1 \\
More than 10 years & 7 \\ \hline
\end{tabular}
\end{center}
\label{experience}
\end{table}%

\begin{table}[htp]
\caption{A breakdown of proficiency levels of participants.}
\begin{center}
\begin{tabular}{| l | r |}
\hline
Proficiency & Count \\ \hline
Not Proficient Whatsoever & 0 \\
Minimal Proficiency & 2 \\
Some Proficiency & 2 \\
Proficient & 8 \\ 
Very Proficient & 10 \\
Extremely Proficient & 1 \\
\hline
\end{tabular}
\end{center}
\label{proficiency}
\end{table}%

Respondents were asked how they had heard about the survey. This questions served as a means to determine which platform attracted the most participants. The data is useful for future surveys, however, bears no effect on the analysis of the results of this survey. Five (N = 5; 22\%) of the participants recruited via email have an existing professional relationship with one the authors.

\begin{table}[htp]
\caption{A breakdown of how participants heard about the survey.}
\begin{center}
\begin{tabular}{| l | r |}
\hline
Source & Count \\ \hline
Reddit & 1 \\
Twitter & 6 \\ 
LinkedIn & 5 \\
Facebook & 4 \\
Email & 5 \\
Referal & 2 \\ \hline

\end{tabular}
\end{center}
\label{source}
\end{table}%

An additional question to determine how familiar respondents were with the field prior to answering was posed. To do this, some common and other, more niche terms were selected by the authors to gauge that prior knowledge. The question listed terminology associated with evolutionary algorithms. The most frequent terms that participants had previously encountered were ``genetic algorithm" and ``evolutionary programming." Both had been selected 15 times. The least frequent terms were ``selective search" and ``adjoint state method." These terms were selected 4 times and 3 times, respectively. One (N = 1; 4\%) participant did not select any terms and none of the participants selected all the terms.

\begin{table}[htp]
\caption{A list of terms that were presented to participants and the frequency with which they had heard of the terms.}
\begin{center}
\begin{tabular}{|c|c|}
\hline
Terminology & Count \\ \hline
Genetic algorithm & 15 \\
Heuristic search algorithm & 10 \\
Evolutionary programming & 15 \\
Selective search & 4 \\
Gradient descent & 11 \\
Simulated annealing & 9 \\
Adjoint state method & 3 \\ \hline
\end{tabular}
\end{center}
\label{terms}
\end{table}%

\subsection{Thematic Analysis}

The technique used to evaluate the results of the open-ended questions in this survey is thematic analysis. The process behind the evaluation is based on the work of Braun \& Clarke \cite{braun2006using}. The steps outlined have become standard practice to conduct similar research in areas such as psychology, sports science, and engineering \cite{terry2017thematic, braun2016using, memarian2018exploring}. A brief overview of each step is given.

\begin{enumerate}
\item Initially, a general familiarisation of the data is conducted to get a general overview of the contents. A rough idea of codes is written down and used as a starting point for the next step.
\item Once the data has been familiarised, the process of coding begins, which is the labelling the data that appears interesting. This can be one or two words within the data or a short segment that conveys a narrow idea.
\item Once the data has been coded, the codes themselves are grouped together in general themes. The relationship between themes is also explored at this stage.
\item Themes are reviewed at this point. Some themes may need to be split, combined or otherwise adjusted. Importantly, a pattern must be evident within the theme among the codes. A complete review of the data is also done at this point to bridge any gaps that may have been missed initially.
\item Themes are concretely defined at this step. Again, themes may be restructured to ensure that each theme has one central idea.
\item Lastly, the analysis of the data is written up. The analysis is broken into individual themes using codes and data to support the arguments.

\end{enumerate}

Three metrics to ensure the highest quality of analysis are outlined by Nowell \emph{et al.} \cite{nowell2017thematic} and consist of credibility, transferability and dependability. Through the realisation of the three metrics, then a fourth metric is fulfilled according to Guba \& Lincoln \cite{guba1989fourth}: conformability. To ensure that these metrics are being met, Nowell \emph{et al.} \cite{nowell2017thematic} suggest an audit trail is set up so that any researcher could follow the logic. Also suggested is that researchers remain critical of their own findings. Electronic records of the analysis were shared among the authors.

Acknowledging internal bias is key to maintaining credibility. In an update on their 2006 paper, Braun \& Clarke \cite{braun2019reflecting} describe researcher bias as an inherent component to thematic analysis and point out that it is the researchers who generate the themes rather than the data itself. Knowledge of one's own bias is 1 of 9 recommendations provided by Castleberry \& Nolen \cite{castleberry2018thematic}. As per the recommendation, bias is discussed later.

\section{Results}

\subsection{Rating Criteria for Design Tools}

This part contained five questions which assessed the respondents' preferences with regards to engineering design tools. The criteria are user interface, versatility, robustness, frequency of use in industry, and supporting documentation. The questions were evaluated on a 3-point Likert scale; the participants had the option to choose from ``no importance," ``some importance" and ``high importance." An additional option of ``no opinion" was provided for respondents.

\begin{figure}[htbp]
\begin{center}
\includegraphics[width=\linewidth]{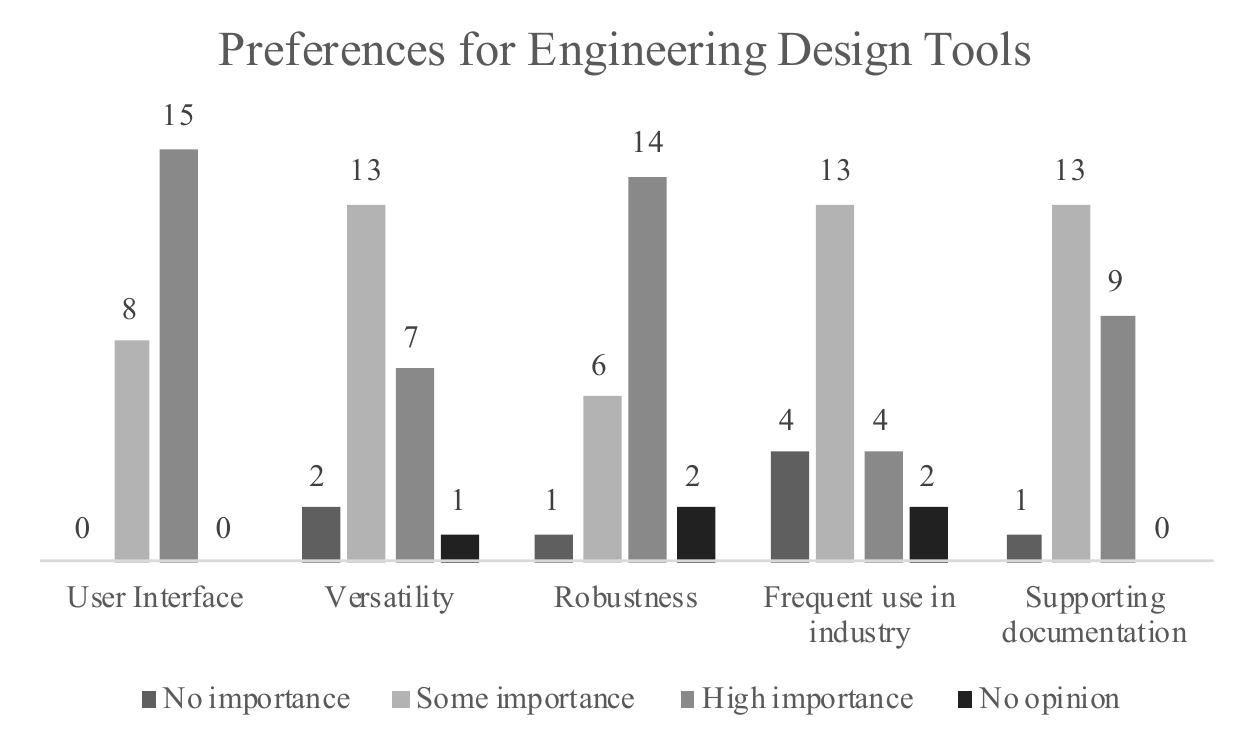}
\caption{User preferences of 5 unique design tool criteria, ranked on a 3-point Likert scale with an additional option of "no opinion."}
\label{fig:vinca1}
\end{center}
\end{figure}


As seen in Figure \ref{fig:vinca1}, the two most important criteria according to participants was user interface (UI) and robustness (Ro). Both of these criteria had a majority (UI: N = 15; 65\%, Ro: N = 14; 61\%) of respondents mark them as highly important. With respect to UI, the participants declared that it is at least somewhat important when considering design tools. From the answers given by participants, the criteria can be ranked: user interface, robustness, supporting documentation, versatility, and frequency of use in industry. This ranking however does not negate the need for any of the criteria as all had been ranked as somewhat important by the majority of participants.

\begin{figure}[htbp]
\begin{center}
\includegraphics[width=\linewidth]{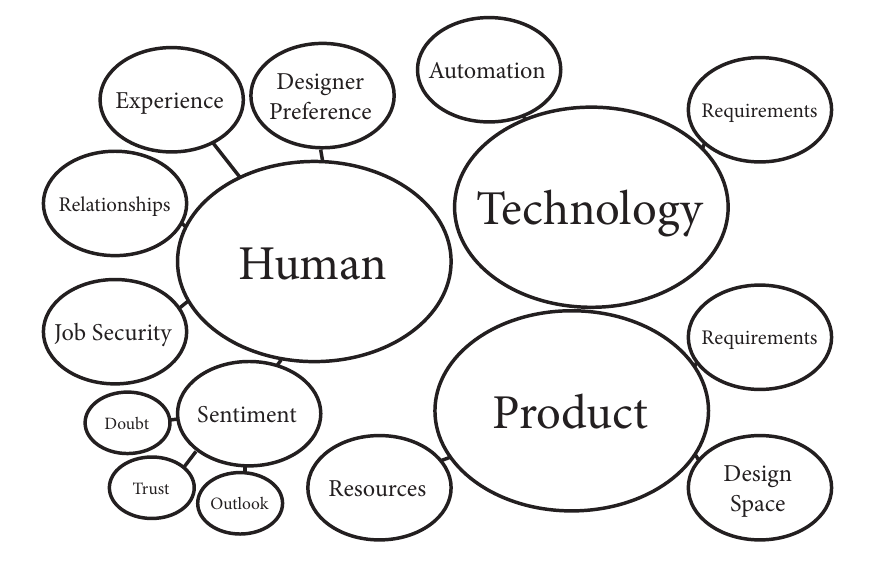}
\caption{A mind map of the themes developed based on the survey data. The nodes emerging from the three main themes (human, product, technology) are the codes which were generated from participants' answers.}
\label{fig:vinca4}
\end{center}
\end{figure}

From the thematic analysis, 3 themes were generated and were given broad names to serve as an overarching link: \emph{human}, \emph{product}, and \emph{technology}. The associated codes are visualised in Figure \ref{fig:vinca4}.

\subsection{Open-Ended Questions}
\subsubsection{Question 1: Describe your current process for optimizing designs; include proportion of time spent on each stage if possible.}

A total of 16 participants answered the first question. Coding could be done on 12 of the answers. The remaining 4 answers contained insufficient detail to accurately assess the meaning and intention of the respondent, or did not answer the question in a coherent manner. The incoherency of answers in this question and following questions had to do with a lack of any sentence structure. The first open-ended question was geared towards answering the second research question.

When discussing the current state of their design processes, participants approached the question from different angles. Using the generated themes as a starting point, 4 included a \emph{human} element, 7 from a \emph{product} perspective and 3 made reference to \emph{technology}. 2 answers had two themes present. Many of the answers that fell under the \emph{product} theme were heavily focused on the design itself. This included talking about the requirements of the design, refinement and optimisation. Participants were also asked to included proportions of their time spent at each stage of the design process. 3 participants gave a breakdown while 1 participant only noted the longest stage of the process. For these 4 participants, iteration/optimisation/redesign takes a significant portion of their time. Some participants also included other considerations into the design process such as designer experience, designer preferences and customer/supplier relationships. With regards to customer requirements, participants said that those requirements are used as a starting point for high-level, initial designs. Combined with answers that gave design process time breakdowns, this stage takes up about 25\% to 40\% of the process. 

One participant, a manager, also noted that finding an optimal solution is ``highly unlikely" although admitted that solutions could be found that are quite close to the optimum. This is in line with another participant, an engineer, who mentioned Multidisciplinary Optimisation (MDO) which invariably results in compromises. These in turn lead to less-than-optimal designs. Of course, this is acceptable as long as the design requirements are met.

\subsubsection{Question 2: What barriers do you have to overcome during the design process?}

This questions was answered by 16 participants. Again, this question was related to the second research question.

Most (N = 11; 69\%) of the participants' answers in the second question fell under the \emph{product} theme. 5 participants explicitly stated that cost was a barrier to their design process. Other resources, such as time and processing power were also mentioned. Previously, it was noted that some participants use the customer requirements as a starting point for their initial design. For one participant, this was incidentally their biggest obstacle. This lends itself to another problem, where the customer does not have clear requirements. Both optimisation algorithms and engineers rely on a clearly defined objective around which to construct their solutions.

One engineer relayed that selecting weights for a fitness function was a barrier to their design process. This is indicative of an over-reliance on the expertise of the engineer and a lack of technology meeting its potential. Spending extra time determining weights takes time away from engineers adding value in their area of expertise. 

Both a student and an engineer mentioned competing design requirements as a barrier. Incorporating conflicting requirements can be challenging, especially when, as mentioned earlier, requirements can be unclear or ill-defined.

Another two participants, a student and a design engineer, mentioned their own knowledge as a limitation to the design process. For the student, being confident enough to be able to get their design from a mental model to a CAD model was a hurdle. For the engineer, the ``learning curve of toolsets" was one of the obstacles. These two answers offer insight into which aspects of the design process can be improved with (better) technology and which are subject purely to engineers. 

Two instances of trust were brought up: one from an engineer and another from a manager. The first had to do with trust in the technology itself, specifically design simulations. The results from simulations must be accurate so that engineers can make informed decisions about the design and ultimately the final product. The second instance of trust was related to the participant's peers. Getting buy-in from new colleagues was mentioned as a barrier. For this participant, trust was instilled as a result of the outcome of the project. This is one manner in which GAs can gain an engineer's trust, although it would still require a high level of trust from key early adopters.

\subsubsection{Question 3: What comes to mind when you think of evolutionary algorithms?}

This question was answered by 16 participants. The intention of this question was to gauge whether respondents have any prejudices or preferences towards algorithms as a part of the design process. This question was in line with the first research question.

The answers in this question came under two themes: \emph{technology} and \emph{human}. The count for each was 10 and 2 respectively. 4 answers were not coded: of these, 3 pertained to biology and did not fall in among the themes while 1 answer was too short to accurately determine the respondents intention. The answers which related to biology were not put into a theme because the creation of an additional theme did not seem appropriate for 3 out of 101 total answers.

Two managers expressed a negative sentiment towards the term used in the question: evolutionary algorithm. The first participant stated that it is ``over-rated in practice" and preferred more rudimentary optimisation methods; pattern search was given as an example. The second participant questioned the ability for evolutionary algorithms to add value. Both participants point to a hurdle that new technology faces: convincing stakeholders that learning a new technology or system is worth the time investment. This sentiment was echoed by an engineer when responding to the last open-ended question. By noting the shortcomings of the current state of GAs, a clear pathway to wider adoption can be determined by addressing their concerns. 

Other participants related the evolutionary algorithm to an automated processes or global optimisation. Three different participants mentioned Artificial Intelligence (AI) and Machine Learning (ML). Seven participants used a variation of selective process, biology and survival of the fittest in their answers. These answers indicate that developers of these algorithms should include a succinct explanation to let engineers have a better understanding of their mechanisms.

\subsubsection{Question 4: Do you trust the designs produced by automated optimisation algorithms? Explain your answer.}

This question was answered by 18 participants. As with the previous question, it was aimed at answering the first research question. 

Participants could very easily be categorised in this question. 13 participants expressed varying sentiment of doubt regarding the result produced by an algorithm. As an example, participants noted that they prefer to check the results of an algorithm by hand. One design engineer extended this by stating that the results would have to be validated by a real world test. This is an example of maintaining the human element in the design process. Keeping the human element in the process is threaded in nearly every answer. Another design engineer noted that ``there are some processes in optimisation that require experience and intuition," which is a direct call to designer expertise. Likewise, one student incorporated the entire lifecycle of the product in their answer; cost, manufacturing and feasibility for human use. A similar sentiment was shared by another student in that some results may be outside of the constraints of the design space. 

Three participants stated an outright trust in results generated by algorithms. One of these respondents, an engineer, states that the results are ``usually overchecked." This is contrary to another engineer's answer, who stated ``a `blind' trust [in automated optimisation algorithms] is a bad approach." Both sides point to the need for a proper explanation of the potential applications of these algorithms as well as the limitations.

This question also revealed some feelings that designers have towards optimisation algorithms. Designers will not trust the final design unless they can understand the process the algorithm took to yield that result.

\subsubsection{Question 5: Do you think a computer-based algorithm could help your design process? Explain your answer.}

This question was answered by 18 participants, although one answer can be discarded as the respondent simply stated that the question was not applicable. This question was related to both research questions.

Two thirds (N = 12; 67\%) of participants expressed a positive sentiment towards working with algorithms in their design process. Much like the previous question, the degree to which participants wanted to incorporate this technology varied. One manager stated that they have already incorporated optimisations into their process. A student declared that they would use it only for the initial design stage. The development of this technology should be able to cater to those who want to use it for one or all parts of during their design processes.

One design engineer gave requirements in anticipation of such technology being incorporated in their process. The most important requirements for this participant was a properly designed UI and a link to other engineering tools. A seamless experience allows engineers to focus on designing rather than debugging. It also avoids user frustration.

\subsubsection{Question 6: Do you have any reservations about implementing more computer-based assistance in the design process? Explain your answer.}

This question asked participants if they have any reservations with regards to implementing more optimisation algorithms in their design processes. It was related to the first research question and was answered by 17 participants.

Each answer could be attributed to the \emph{human} theme, though through various degrees. One student drew comparisons between other software such as CAD and Computational Fluid Dynamics, noting that the design process is already heavily computer based.

An engineer and a student are concerned about the reliance that future engineers and designers will have on systems like GAs. The engineer mentioned that these algorithms have ``disengaged the brains of engineers" which threatens a long-term consequence of engineers that have poor design experience.

Two participants, a student and an engineer, cited job security as a reason why they would not want this technology implemented. This is a very real concern that needs to be addressed by properly explaining the intent of developing these tools.

Another engineer was very supportive of the idea, mentioning that they have been advocating for this kind of technology for the last 20 years. It is an accurate reflection of the challenge of getting new technology to be adopted by individuals and by organisations alike.

\subsubsection{Open-Ended Question Summuray}

\begin{figure}[htbp]
\begin{center}
\includegraphics[width=\linewidth]{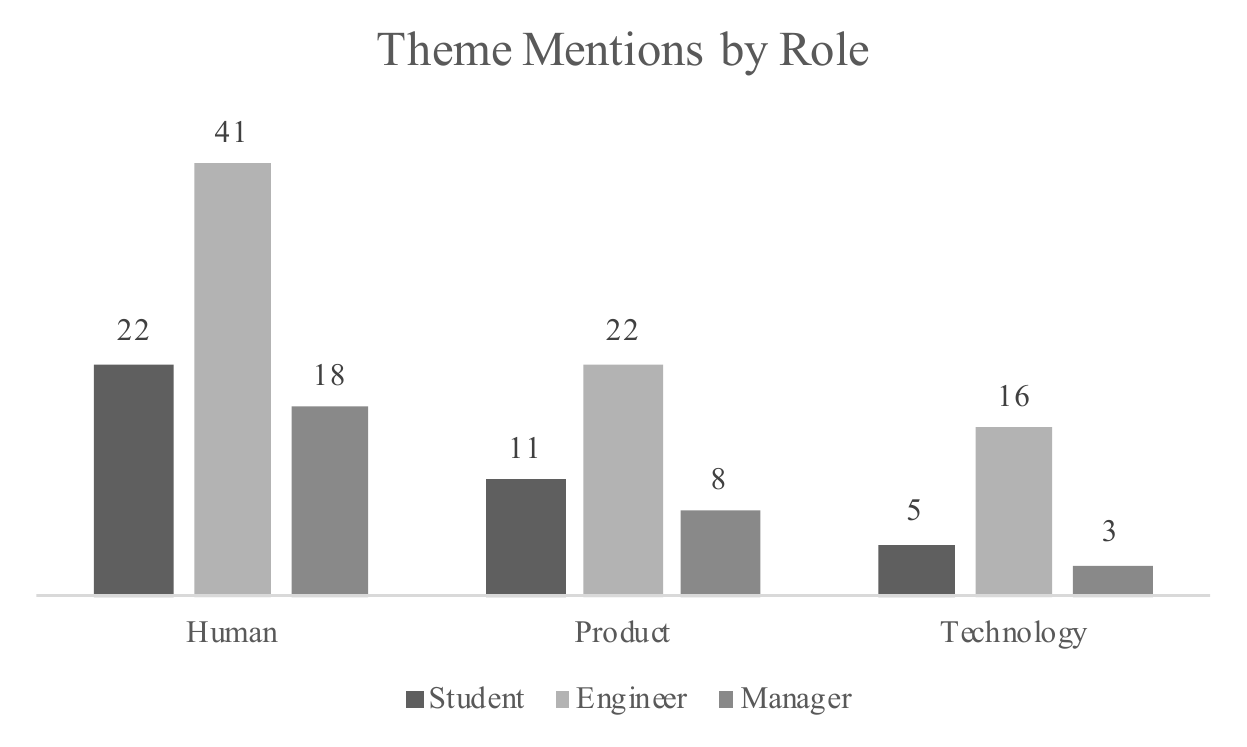}
\caption{The raw count of mentions per theme by each different role in all questions.}
\label{fig:vinca2}
\end{center}
\end{figure}

\begin{figure}[htbp]
\begin{center}
\includegraphics[width=\linewidth]{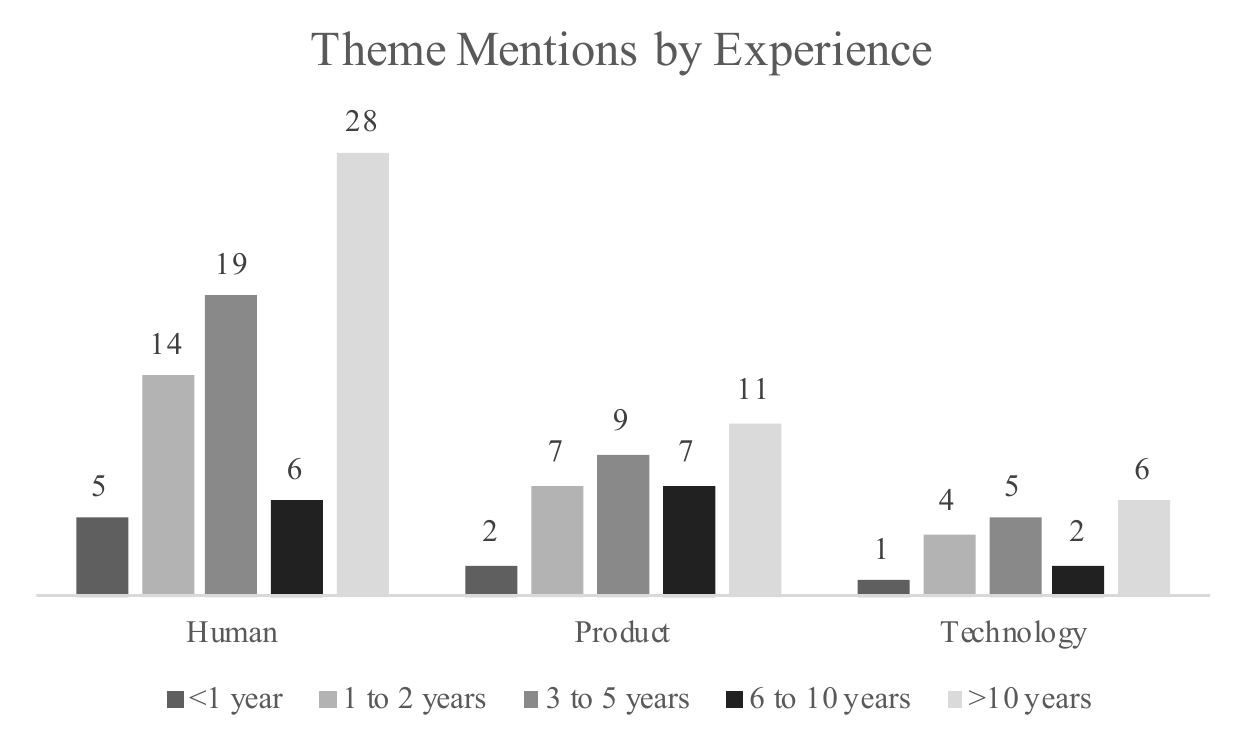}
\caption{The raw count of mentions per theme by each different experience level in all questions.}
\label{fig:vinca3}
\end{center}
\end{figure}

The amount of mentions per theme are summarised in Figure \ref{fig:vinca2} and Figure \ref{fig:vinca3}. Discrepencies in theme mentions can be explained by a varying number of participants in each category. The number of mentions also exceeds the total number of answers as some answers contained more than one theme.

\section{Discussion}

A common thread throughout participants responses is that there is a question of trust towards GAs within industry. Perhaps surprising is that the key to gaining this trust is not producing good results, but creating algorithms which explain the process they take in reaching a result. Participants have expressed a desire to continue to remain in the design loop. This is at odds with the motivation of a portion of the GA community of removing humans from the loop \cite{shahriari2015taking}. It is clear we need to take a different approach to increase industrial uptake.

The participants in this study have demonstrated through their answers that there is a general distrust towards GA-based design in industry. 72\% of answers expressed doubt when discussing whether designs produced by an optimisation algorithm could be trusted. Their reservations have to do with the unproven state of these algorithms in their own personal experiences. However, respondents also recognise the potential of this technology, with 76\% of respondents saying they could see the value of integrating GAs somewhere in their process. Of the respondents that expressed doubt, 69\% of them noted that these algorithms could also help their design process. Separately, participants noted that the inclusion of human expertise is vital for adoption.

Much to the surprise of the authors, the answers between the different groups (students, engineers, managers) and experience levels did not vary significantly. This can be seen in Figure \ref{fig:vinca2} and Figure \ref{fig:vinca3}. The different groups shared the same thoughts across a variety of questions, especially when considering the two research questions. The most notable example of this is that both a student and an engineer were concerned about their jobs being replaced or made redundant by automation.  

Optimisation and iterative design are significant portions of the design process. When discussing their design process and the time spent at each stage, one participant said that in their experience ``the detailed subsystem design is usually the longest stage" and that ``an iterative evolutionary algorithm would speed up this process." This is in conjunction with the previous statement that this stage takes up 25\% to 40\% of the total design process time.

Recognising that trust is important in the development of new algorithms and the integration of such algorithms in industry is key to their adoption. Participants noted that trust plays a big role in the adoption of new technologies. One participant had a clear mistrust of new algorithms. Others mention multiple times that the need to be able to review the results and the decisions that an algorithm takes is a priority. When asked whether they had any reservations towards optimisation algorithms, three participants explicitly said yes. One of these participants went on to ask a series of questions about the ethics of the decisions made by an algorithm. Questions included ``Who would be liable for a death or damage caused by a design made by this algorithm?" and ``Are we going to write a decisions rules book?" which clearly demonstrates that designers are not only aware of the decisions they make with respect to their design, but also to the wider consequences.

The sophistication of the design process was also different across the range of participants. Two participants had a more traditional approach when refining their design, opting for ``trial and error" and ``one factor at a time." These approaches are useful for exploring the entire design space and are a mix between designer experience and technology. However, they can be very time-consuming processes, especially on complex designs. While a Design of Experiments (DoE) is generally suitable for proof that a global optimum has been achieved, a trial and error method is not.

\section{Recommendations}

The participants in this study have shown with their answers that there is one key element which will determine the scale of adoption - trust. This principle is used as a foundation for the following recommendations.

\subsection{Algorithm Developers Should Trust Designers and Engineers}
By trusting designers to know what's best for them, developers of new GAs should engage with experts in creating solutions aimed at solving their optimisation problems. While the development of new algorithms is important in addressing different optimisation challenges, the application must go beyond benchmark tests and into industry for that algorithm to realise its potential. Moreover, the reservations that participants had towards these algorithms point to a general need for algorithms to have suitable explainability and transparency. These criteria correspond closely to the requirements set out by those researching the ethics and law of AI \cite{eu2020white}. The domain of evolutionary algorithms could coopt the proposed regulatory framework of AI as a starting point as these technologies become more prevalent in industry. 

\subsection{Algorithm Developers Should Increase Involvement of Human-Computer Interaction Research}
Throughout the survey, the need for human involvement in the development process of GAs is brought up; it demonstrates a need for a user-centric approach. Human-Computer Interaction (HCI) is the field of study which concerns itself with researching the relationship between users and digital technology. This can first be observed in Part 2 of the survey, where respondents identified a good UI as the most important factor when considering a design tool. The expectations from designers is that any additional tools that are made must integrate into their existing design process. This is articulated by a participant who needs an ``intuitive set-up and an interface to existing tools." This can be seen as a direct answer to the second research question which enquires about the design requirements of domain experts. These tools need to also cater to novice users such as students. A user-friendly interface can lower the barrier of entry to GAs. An engagement between HCI researchers and GA researchers can further these algorithms towards a higher adoption rate in industry.

\subsection{Algorithm Developers Should Develop Better Communication Links Between Researchers and Engineers}
The need for solutions to be developed in conjunction with the designers who will eventually be using these tools is evident in the answers of participants. Integrating constraints in the design optimisation process was a recurrent subject among participants' answers. Simplifying the problem down to something that could be modelled and optimised while maintaining a feasible enough design is a challenge one of the participants cited. The human aspect of design was also brought up by the participants by referencing the necessity for designer expertise in more complex designs. One of the participants also talked about aesthetic constraints. This is a constraint that cannot be modelled and relies solely on the experience and preferences of the designer. Moreover, it should also serve as a reminder that any engineering design tool developed is not meant to cover every single aspect of the design process. The expectations of engineers needs to be set by researchers who develop these tools to determine what constitutes ``state-of-the-art" with respect to this technology and what the limitations are of these techniques. Analogously, GAs are not the Swiss Army Knife of design tools.

\subsection{Algorithm Developers Should Include Visualisation as a Part of Their Development Process}
Engaging with designers and building their trust in GAs can be done with the help of visualisations. A key aspect of building trust is to develop a level of understanding of the algorithms. A simple manner in demonstrating the performance of an algorithm is to visualise the resulting solution after each generation. This is especially useful when comparing two or more algorithms. If the algorithms themselves are considered as tools, then this would give designers an easy manner to comprehend which algorithm is better for their application. For some designers, it would also be beneficial to see the different solutions that a GA can produce. This can be especially useful for designers who also need to consider aesthetic constraints. It also leads to another consideration in the development of GAs. The visualisation of algorithms is mentioned in the Introduction section with work involving construction site layouts \cite{kaveh2016construction}.

\section{Evaluation of Our Approach}

As recommended by Castleberry \& Nolen \cite{castleberry2018thematic}, personal bias is addressed. Every effort has been made to mitigate the effects of bias when analysing, presenting, and discussing the data. With regards to pre-survey bias, the design of the study along with the questions were reviewed by the authors and amended to keep the study as objective as possible. As this survey was done entirely online, interview bias was not present. During the entire duration of the survey being available to respondents, the answers were not gleaned to prevent any bias during the analysis stage. Of the 23 respondents that took part in this study, 7 could definitively be attributed as a professional connection to one of the authors (through answers provided to the question ``How did you hear about this survey?") which introduces some level of bias. This was minimised by not asking respondents for any identifiable and traceable details. Moreover, not a single answer contained any identifiable words or phrases.

Under a restriction-free scenario, participants for this study would have been interviewed in-person; due to COVID-19, this could not be done. This did mean that the participants could not be prompted further when discussing their answers. As an example, some of the answers could not be coded due to their brevity so further detail from respondents would have been beneficial. In the future, video conferencing could be used instead of face-to-face interviews.

\section{Summary}

A total of 23 participants were recruited through digital means to answer a survey. Thematic analysis was performed on 6 open-ended questions, which yielded 3 major themes: \emph{human}, \emph{technology}, \emph{product}. With regards to their current design processes, respondents noted that optimisation takes a significant portion of their time. Respondents were generally aware of the source of inspiration from which evolutionary algorithms take their names, but lacked an in-depth knowledge of the subject. A majority of participants stated that they would implement these algorithms in their design process, although some still held reservations about their implementation.

With respect to the two research questions posed, participants have demonstrated through their answers that there is a general distrust towards GA-based design, yet there is also an acknowledgement among participants that these algorithms could benefit the engineering design process. Trust in these algorithms was one of the barriers to adoption. Cost, manufacturability, and knowledge limitations were also mentioned as hurdles in the design engineering process. 

Practitioners of GA tools want to understand how the algorithm made its decision. Making tools for designers that are intuitive and whose functionality can be accessed easily by engineers is paramount; future GA interfaces should maximise engagement with designers. A clear communication channel between researchers and their end users can increase the adoption rate. Trust is an important factor for designers. By allowing engineers to be in control of the process, engineers will be able to build a level of trust with the design tools. Likewise, a human-centred approach will demonstrate to designers that these tools are not there to replace them.

\section*{Acknowledgment}

J.V. would like to thank members of the EPIC CDT for their continuous support.

\ifCLASSOPTIONcaptionsoff
  \newpage
\fi



\bibliographystyle{IEEEtran}
%
\bibliography{EDObib}



%








\end{document}